\documentclass[letterpaper]{article} % DO NOT CHANGE THIS
\usepackage{aaai2026}  % DO NOT CHANGE THIS
\usepackage{booktabs}
\usepackage{times}  % DO NOT CHANGE THIS
\usepackage{helvet}  % DO NOT CHANGE THIS
\usepackage{courier}  % DO NOT CHANGE THIS
\usepackage[hyphens]{url}  % DO NOT CHANGE THIS
\usepackage{graphicx} % DO NOT CHANGE THIS
\usepackage{natbib}  % DO NOT CHANGE THIS AND DO NOT ADD ANY OPTIONS TO IT
\usepackage{caption} % DO NOT CHANGE THIS AND DO NOT ADD ANY OPTIONS TO IT
\usepackage{subcaption}
\usepackage{algorithm}
\usepackage{algorithmic}
\usepackage{amsmath}

\usepackage{newfloat}
\usepackage{listings}
\usepackage{tcolorbox}
\usepackage{amsfonts}
\usepackage{multirow}
\usepackage{graphicx}

\DeclareCaptionStyle{ruled}{labelfont=normalfont,labelsep=colon,strut=off} % DO NOT CHANGE THIS
\floatstyle{ruled}
\newfloat{listing}{tb}{lst}{}
\floatname{listing}{Listing}
\title{Forecast2Anomaly (F2A): Adapting Multivariate Time Series Foundation Models for Anomaly Prediction}
\author{
    Atif Hassan, 
    Tarun Kumar, 
    Ashish Mishra, 
    Sergey Serebryakov, 
    Satish Kumar Mopur, 
    Phanidhar Koganti, 
    Murthy Chelankuri, 
    Ramanagopal Vogety, 
    Suparna Bhattacharya, 
    Martin Foltin
}
\affiliations{
    Hewlett Packard Enterprise\\
}

\usepackage{bibentry}
\begin{document}

\maketitle

\begin{abstract}
Forecasting anomalies (anomaly prediction) in multivariate time series from different real-world, dynamic and complex systems is vital for preempting critical failures, leading to a substantial minimization in operational costs and human labour. 
Yet, existing methods are limited to specific systems while failing to generalize to evolving anomaly patterns over time.
In contrast, pre-trained Time Series Foundation Models (TSFMs) have recently demonstrated strong generalization and zero-shot forecasting capabilities. 
However, their potential remains untapped for anomaly prediction, a task fundamentally different from forecasting normal behavior.
Thus, we present Forecast2Anomaly (F2A), a novel framework that empowers TSFMs with anomaly prediction abilities through two key innovations. 
First, we propose a joint forecast-anomaly loss that fine-tunes TSFMs to accurately forecast future signals even at anomalous time points. 
Second, we introduce a Retrieval-Augmented  Generation (RAG) module that retrieves historically relevant horizons and conditions predictions on them. 
This component dynamically adapts to distributional shifts at inference time, enabling F2A to track evolving anomalies without requiring model updates.
By combining targeted fine-tuning with dynamic retrieval, F2A bridges the gap between robust TSFM zero-shot forecasting and zero-shot anomaly prediction. 
Extensive experiments across $16$ diverse datasets and multiple TSFM backbones show that F2A consistently outperforms state-of-the-art methods, offering a scalable, zero-shot anomaly prediction solution for real-world applications.
\end{abstract}

\section{Introduction}\label{sec:intro}
Modern digital systems continuously generate vast streams of multivariate time series data such as metrics, logs, and traces that are critical to monitoring system health. 
Traditionally, irregularities in such data are identified using reactive anomaly detection, which detects anomalies only after they manifest~\cite{xu2021anomaly,paffenroth2018robust,malhotra2015long}. 
However, this retrospective approach is inadequate for preventing catastrophic failures, which often carry high operational costs and demand costly human intervention.

Anomaly prediction, which aims to anticipate anomalies before they occur, has emerged as a proactive alternative \cite{parkwill,shyalika2024time}. 
By forecasting the likelihood of future failures, anomaly prediction enables early warning systems, and real-time decision-making. 
However, this task remains underexplored due to its inherent complexity of predicting rare, evolving events from past context alone, without access to future signals. 
The challenge intensifies in multivariate settings, where subtle inter-dependencies can obscure early signs of failure.

A promising direction lies in Time Series Foundation Models (TSFMs), trained on diverse, heterogeneous time series corpora. 
These models demonstrate remarkable generalization for zero-shot forecasting \cite{liang2024foundation, yeh2023toward, goswami2024moment}, making them underexplored but attractive candidates for anomaly prediction in unseen domains.  
Yet, TSFMs suffer from two key limitations in this setting: (i) \textbf{Poor anomaly expressiveness:} TSFMs are pre-trained on predominantly normal signals, leading to smooth forecasts that suppress anomaly signatures (see Fig. \ref{fig:daphnet_vanilla_forecast}), hindering downstream anomaly scoring. (ii) \textbf{Static behavior under distributional shift:} Once trained, TSFMs are frozen and cannot adapt to domain drift or non-stationarity, making them brittle in real-world environments.

To partially address the second issue, Retrieval-Augmented Generation/Forecasting (RAG) has gained traction. 
By retrieving semantically similar past sequences from a database, these methods enable zero-shot forecast adaptation without retraining \cite{hanretrieval, yang2025timerag, liu2024retrieval}.
However, existing time series RAG approaches are not designed for anomaly prediction and suffer from critical flaws. Often methods assume successful temporal alignment between forecast and retrieved horizons (future signal) in the embedding space through a learnable projection layer \cite{ning2025ts} which is a simplistic assumption to make.
On the other hand, multiple techniques often fuse retrieved horizons via static attention mechanisms, ignoring when retrieval helps or hurts \cite{han2025retrieval}. 
In high-confidence predictions, retrieval may inject noise while during difficult or anomalous regions, it may be essential. 
Current methods lack the ability to adaptively modulate this influence.

\subsection{Forecast2Anomaly (F2A): A Unified Framework}
To address the following three gaps, (i) zero-shot anomaly prediction, (ii) poor anomaly-expressiveness in forecasts, and (iii) non-adaptive retrieval fusion, we propose Forecast2Anomaly (F2A), a unified framework that adapts multivariate TSFMs to anomaly prediction.
With F2A, we make the following two key contributions:
\begin{itemize}
    \item \textbf{Joint Forecasting-Anomaly Prediction Training:} We fine-tune the Time Series Foundation Model (TSFM) using a novel multi-task objective that combines both forecasting and anomaly prediction.
    In contrast to prior work~\cite{parkwill}, we adopt the focal loss for anomaly prediction to address the inherent rarity of anomalies. 
    This joint training paradigm encourages the model to produce forecasts that preserve subtle irregularities indicative of impending anomalies, rather than smoothing them out, thereby enhancing predictive sensitivity (see the third section in discussion for details).
    \item \textbf{Anomaly Sensitive Time Series RAG Module:} F2A introduces a specialized retrieval mechanism that retrieves $k$ horizons of the corresponding top-$k$ semantically similar past contexts and fuses them with the model’s forecast through a learned, context-aware aggregation module, 
    bypassing embedding space alignment issues. 
    Since our RAG module is trained jointly to improve forecasting as well as anomaly prediction, it enables selective reliance on external signals only when needed, allowing the model to adapt in zero-shot or non-stationary settings without any gradient updates. 
    % \textcolor{red}{Can you also provide brief justification about how this RAG approach is differ from existing RAG?}
\end{itemize}
As a result, F2A transforms TSFMs from passive forecasters into proactive, retrieval-enhanced early warning systems. 
It is plug-and-play, robust to domain drift, and excels even with lightweight base models in both fine-tuned and zero-shot settings.
We conduct extensive experiments on $16$ datasets from the TSB-AD-M benchmark, including $6$ held-out datasets for zero-shot evaluation. 
Across $3$ distinct TSFMs and against $10$ strong baselines including statistical methods, F2A consistently outperforms in both accuracy and generalization, showcasing the power of retrieval-guided, dual-task TSFMs for real-world anomaly prediction.

\section{Related Works}\label{sec:related-works}
\subsubsection{Anomaly Detection in Time Series:}
Detecting anomalies in multivariate time series is a long-standing problem, traditionally tackled using statistical tests (e.g., control charts, ARIMA, PCA) and classical ML techniques (e.g., One-Class SVM, Isolation Forest). 
These methods struggle in high-dimensional, noisy settings where true anomalies are rare~\cite{zhong2023survey}. 
Recently proposed deep learning approaches such as autoencoders~\cite{sakurada2014anomaly}, LSTMs~\cite{malhotra2015long} and Transformers \cite{xu2021anomaly} improve robustness but largely operate retrospectively, flagging anomalies after they occur~\cite{zamanzadeh2024deep}.

\subsubsection{Time Series Foundation Models (TSFMs):}
TSFMs have emerged as powerful pre-trained models for forecasting and representation learning~\cite{shyalika2024time}. 
Models such as Chronos~\cite{ansarichronos}, TimesFM~\cite{das2024decoder}, Moirai~\cite{woo2024unified}, and TimeGPT~\cite{garza2023timegpt} demonstrate strong zero-shot forecasting across domains. Compact variants like TinyTimeMixer (TTM) achieve competitive performance with fewer than $1$M parameters~\cite{ekambaram2024tiny}. 
While some models like TSPulse~\cite{ekambaram2025tspulse} and Moirai have shown promise in anomaly detection, they are not designed to predict anomalies in advance.

\subsubsection{Retrieval-Augmented Generation for Time Series:}
Originally introduced in NLP~\cite{lewis2020retrieval}, retrieval-augmented generation (RAG) enables models to incorporate external context at inference. 
Recent extensions to time series include TS-RAG~\cite{ning2025ts} and RAFT~\cite{han2025retrieval}, which retrieve semantically similar sequences to boost forecasting. 
These methods improve generalization without retraining, but focus solely on forecasting, not anomaly prediction. 
Moreover, fusion with retrieved signals is often static and lacks task-specific adaptation.

\subsubsection{Anomaly Prediction:}
Anomaly prediction, unlike detection, aims to forecast failures before they occur—making it more suitable for early warning and recovery. 
Anomaly-to-Prompt (A2P)~\cite{parkwill} is a recent method that formulates this task using synthetic anomaly prompts and transformers, but it is not built on TSFMs and therefore lacks their benefits including zero-shot.

\subsubsection{Our Approach:}
F2A differs from prior work in three key ways: (i) it explicitly targets anomaly prediction as a learning objective allowing zero-shot anomaly predictions using TSFMs; (ii) it introduces a dual-task training strategy that refines forecasts to better reflect future anomalies; and (iii) it incorporates a retrieval module that fuses retrieved sequences in an adaptive, anomaly-aware manner. 
To our knowledge, F2A is the first framework to unify TSFMs and RAG for zero-shot anomaly prediction in time series.

% Uncomment the following to link to your code, datasets, an extended version or similar.
% You must keep this block between (not within) the abstract and the main body of the paper.
% \begin{links}
%     \link{Code}{https://aaai.org/example/code}
%     \link{Datasets}{https://aaai.org/example/datasets}
%     \link{Extended version}{https://aaai.org/example/extended-version}
% \end{links}

\section{Preliminaries}\label{sec:methodology}
\subsection{Notations}
Let $\mathcal{X}=\{\mathbf{X}^1,\mathbf{X}^2,\cdots,\mathbf{X}^n\}$ denote a collection of $n$ raw multivariate time series, where each $\mathbf{X}^i \in \mathbb{R}^{T_i \times C_i}$ represents a sequence with $T_i$ time steps and $C_i$ channels (features). 
Given a supervised anomaly prediction task, $Y^i\in\{0,1\}^{T_i}$ denotes the binary labels for each timestep in $\mathbf{X}^i$.
Typically, a sliding window of length $L$ with stride $H$ is applied on the raw sequences $\mathbf{X}^i$ to produce inputs $\widetilde{\mathbf{X}}^i \in \mathbb{R}^{N_i \times L \times C_i}$ and corresponding true future horizons $\mathbf{Z}^i \in \mathbb{R}^{N_i \times H \times C_i}$. 
Here $H$ is the length of the horizon and $N_i=\left\lceil\frac{T_i}{L}\right\rceil$. 
Each window spans $L$ steps of input and is paired with a horizon of $H$ future steps.
Similarly, applying the same sliding window on $Y^i$ results in labels $\mathbf{Y}^i \in \{0,1\}^{N_i \times H}$.
We denote $x\in\widetilde{\mathbf{X}}^i$ as a single input window where $x\in\mathbb{R}^{L\times C_i}$, $y\in\mathbf{Y}^i$ as the corresponding labels where $y\in\{0,1\}^{H}$ and the corresponding true horizon, $z\in\mathbf{Z}^i$ where $z\in\mathbb{R}^{H\times C_i}$.

\subsection{Problem Setup}\label{sec:problem}
We consider the task of anomaly prediction in multivariate time series. 
Given a historical context window of observations, the goal is to (i) forecast future values over a prediction horizon and (ii) proactively identify which future time points are likely to be anomalous.

% Formally, let $\mathcal{X}=\{\mathbf{X}^1,\mathbf{X}^2,\cdots,\mathbf{X}^n\}$ denote a collection of $n$ multivariate time series, where each $\mathbf{X}^i \in \mathbb{R}^{T_i \times C_i}$ represents a sequence with $T_i$ time steps and $C_i$ channels (features). 
For a given sequence $x$, we are interested in predicting anomalies over a fixed number of future timesteps, $H$.
To this end, we define,
\begin{itemize}
    \item A \textbf{forecasting function}, $f_{\mathrm{f}}: x \rightarrow \widehat{x}$ where $\widehat{x}$ is a multivariate forecast such that $\widehat{x}\in \mathbb{R}^{H \times C_i}$.
    \item An \textbf{anomaly prediction function}, $f_{\mathrm{ap}}: \widehat{x} \rightarrow p$, which produces a probability vector $p \in \mathbb{R}^{H}$, where each element $p_t$ represents the predicted probability of the $t$-th future timestep being anomalous.
    To obtain binary anomaly labels, we apply a threshold $u$, such that,
    \begin{equation}
        y_t=\begin{cases}
            1 ~~~\text{ if }~~ p_t>u\\
            0 ~~~\text{ otherwise}
        \end{cases}
    \end{equation}
    Here $u$ is a threshold and $y_t = 1$ indicates that the $t$-th timestep is predicted to be anomalous.
\end{itemize}
Thus, the goal of anomaly prediction is to learn $f_{\mathrm{ap}}$ conditioned on $f_{\mathrm{f}}$ without having access to the true future $z$ (hence a harder task than anomaly detection). 

\section{Methodology}\label{sec:proposed-solution}
We propose, F2A, a framework that adapts any pre-trained TSFM to predict anomalies by fine-tuning the model using a novel loss function while fusing contextual information from a novel, learnable retrieval mechanism.

% \subsection{Pre-processing}
% \subsubsection{Data Preparation:}
% Given raw multivariate time series $\mathbf{X}^i \in \mathbb{R}^{T_i\times C_i}$ having $T_i$ number of time steps and $C_i$ number of channels, we apply a sliding window of length $L$ with stride $H$ to produce inputs $\widetilde{\mathbf{X}}^i \in \mathbb{R}^{N_i \times L \times C_i}$, labels $\mathbf{Y}^i \in \{0,1\}^{N_i \times H}$, and true future horizons $\mathbf{Z}^i \in \mathbb{R}^{N_i \times H \times C_i}$, where $H$ is the prediction horizon and $N_i=\left\lceil\frac{T_i}{L}\right\rceil$.

\subsection{Channel Selection (Pre-processing)} 
Multivariate time series often have varying channel counts ($C_i$), but modern deep learning training pipelines require a fixed input dimensionality. 
To standardize inputs, we select a fixed number of channels $C$ across all samples. 
We rank channels by the variance of their first-order differenced signals, highlighting sharp transitions typical of anomalies while suppressing noise and trends. 
The top $C$ channels are retained with zero-padding being applied if $C_i < C$.

\subsection{Joint Anomaly-Forecast Loss}
Anomaly prediction is a highly imbalanced binary classification task, where the model must estimate the probability of an anomaly at each future timestep.
Crucially, the quality of these predictions depends on the forecast being expressive enough to retain signatures of anomalous behavior.
This is challenging when using time series foundation models (TSFMs) as forecasters.
While TSFMs generalize well across domains, they are typically pre-trained on large corpora where anomalies are rare and without explicit objectives to capture anomaly patterns.
As a result, they tend to produce overly smoothed forecasts that suppress irregular or extreme deviations, impairing downstream anomaly prediction.
To address this, we propose a joint training objective that optimizes both the anomaly predictor, $f_{\mathrm{ap}}$, and the forecaster $f_{\mathrm{f}}$ for improved predictive accuracy.
Given a training triple $(x, y, z)$, where $x$ is the multivariate input window, $y$ the ground truth anomaly label, and $z$ the future multivariate signal (horizon), we define the loss function $L$ as:
\begin{equation}
    \mathcal{L} = \sum_{t=1}^H\mathcal{L}_{\mathrm{ap}}(p_t, y_t) + \lambda\underbrace{ \frac{1}{H}\sum_{t=1}^Hm_t\left\lvert
\widehat{x}_t- z_t\right\rvert}_{L_f}
\label{eqn:loss_function}
\end{equation}
Here, $p_t = f_{\mathrm{ap}}(x)_t$ is the predicted probability of an anomaly at timestep $t$, $\widehat{x}_t = f_{\mathrm{f}}(x)_t \in \mathbb{R}^C$ is the predicted multivariate forecast at timestep $t$, $\mathcal{L}_\mathrm{ap}$ is the focal loss \cite{lin2017focal} to handle class imbalance, $\lambda$ is a weighting hyperparameter to balance $\mathcal{L}_\mathrm{ap}$ and $\mathcal{L}_\mathrm{f}$ and $m_t$ is a timestep-specific weighting term defined as:
    \begin{equation}
    m_t=\begin{cases}
        1~~~\text{ if }~y_t=0\\
        \psi~~~\text{ if }~y_t=1\\
    \end{cases}
\end{equation}
where, typically, $\psi\geq1$.
Eqn. \ref{eqn:loss_function} essentially combines two separate losses, a classification loss $\mathcal{L}_{\text{ap}}$ and a forecasting loss $\mathcal{L}_f$. 
We use the focal loss \cite{lin2017focal} as $\mathcal{L}_{\text{ap}}$ since it  encourages confident anomaly predictions in an imbalanced setting, while $\mathcal{L}_f$ is a weighted mean absolute error which guides the forecaster to prioritize accurate prediction of anomalous regions in the signal by upweighting errors where anomalies are labeled. 
Note that the ground truth future signal, $z$, is used only during training and not inference.

By jointly optimizing this loss, the model learns to generate forecasts that preserve anomaly-indicative features, enabling more effective and interpretable anomaly prediction. Notably, our anomaly-aware forecasting loss is agnostic to the base TSFM architecture and can be applied on top of any pretrained forecaster, making it a plug-and-play enhancement module for TSFM-based anomaly prediction pipelines.

\begin{figure*}[t]
    \centering
    \includegraphics[width=0.9\linewidth]{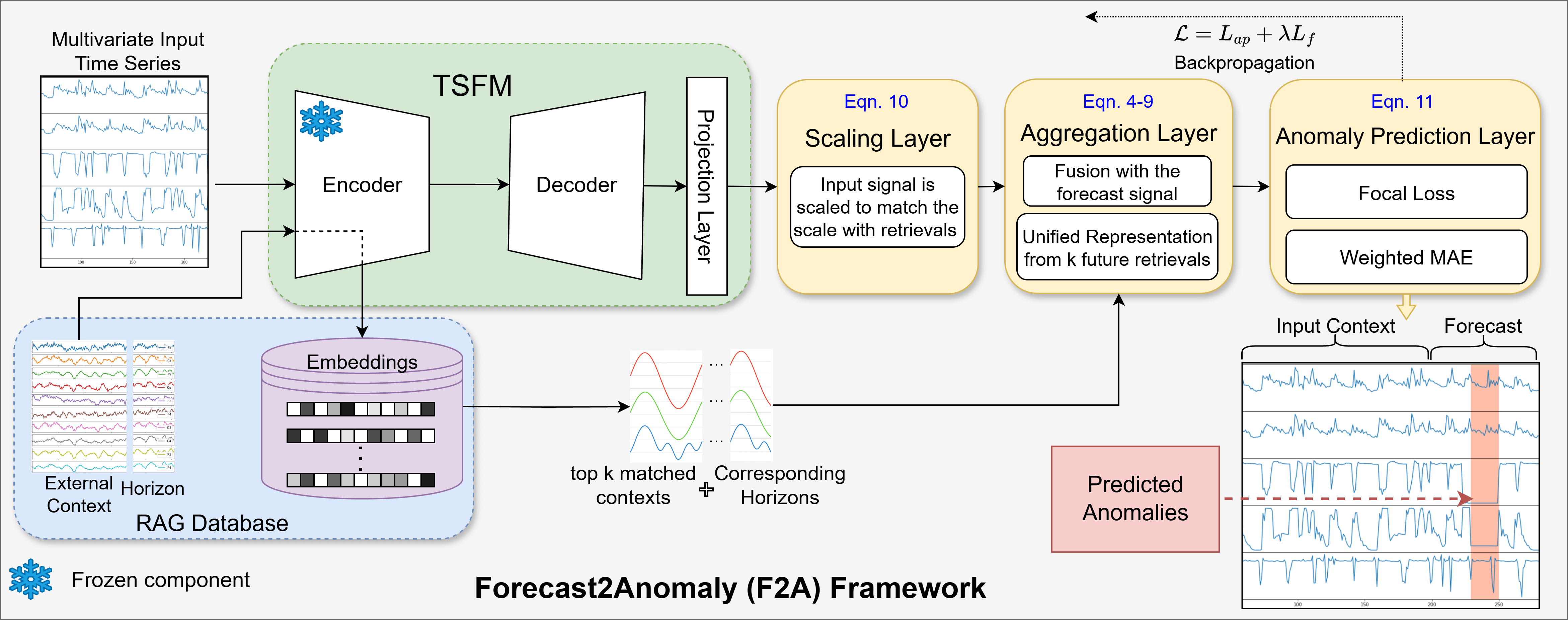}
    \caption{Overview of our proposed framework, F2A. 
    We assume that the TSFM has an encoder-decoder architecture wherein we freeze the encoder while all other parameters are fine-tuned.}
    \label{fig:f2a}
\end{figure*}

\subsection{Retrieval-Augmented Generation (RAG)}
Our RAG architecture comprises a retrieval module and an aggregation layer. 
The goal is to augment the base forecast with relevant past signals retrieved from a database, enabling more expressive and anomaly-aware predictions.

\subsubsection{Retrieval:}
To construct the retrieval database, we curate a collection $\mathcal{B}$ of triplets, $\mathcal{B}=\{(x^i,e^i,z^i)\}_{i=1}^{\lvert\mathcal{B}\rvert}$ where $x^i\in\mathbb{R}^{L\times C}$ is the multivariate input window, $e^i\in\mathbb{R}^{C\times D}$ is its corresponding embedding, typically derived from a frozen copy of the encoder of the forecasting model, $f_{\mathrm{f}}$, where $D$ is the dimension of the encoder representation and $z^i\in\mathbb{R}^{H\times C}$ is the corresponding future signal. 
For vector similarity search, each embedding $e^i$ is flattened to $\mathbb{R}^{CD}$, preserving channel-wise temporal structure.
Given an input signal, $x^j$, its corresponding embedding $e^j$ is derived and $k$ future signals, $\{z^q\}_{q=1}^k\subset\mathcal{B}$ are retrieved such that $\forall~i\in\{1,2,\cdots,\left\lvert\mathcal{B}\right\rvert\}$, the $\ell^2$ distance between $e^j$ and $e^i$ are minimized. 
Let $o = (z^1, z^2, \ldots, z^k) \in \mathbb{R}^{k \times H \times C}$ denote the retrieved future signals. 
These sequences, along with the base model’s forecast $\widehat{x}^j = f_{\mathrm{f}}(x^j)$, are passed to an aggregation layer designed to produce an enhanced final forecast.

\subsubsection{Aggregation Layer:}
The first aggregation step synthesizes information across the $k$ retrieved futures to produce a unified representation:
    \begin{align}
        \widehat{o} &= \tau(o)\label{eqn:reshape1}\\
        \phi &= \mathrm{softmax}\left(\widehat{o}W^1\right)\label{eqn:weights1}\\
        \widetilde{o} &= \tau^{-1}\left(\phi\widehat{o}\right)\label{eqn:inv_reshape1}\\
        h^1&=\sum_{r=1}^k\widetilde{o}^r\label{eqn:final_rep1}
    \end{align}
Eqn. \ref{eqn:reshape1} reshapes the concatenated future signals along the temporal axis to yield $\widehat{o}\in\mathbb{R}^{kH\times C}$
Here, $\tau(\cdot)$ is the reshaping function.
Next, a set of weights for each timestep across all signals are generated using Eqn. \ref{eqn:weights1} where $W^1\in\mathbb{R}^{C\times 1}$ is a learnable parameter and $\phi\in\mathbb{R}^{kH\times 1}$. 
These learned importance scores are used to weigh each timestep as shown in Eqn. \ref{eqn:inv_reshape1} with $\tau^{-1}(\cdot)$ reshaping its input back to the original shape. 
Finally, the reweighted timesteps of individual future signals are summed as shown in Eqn. \ref{eqn:final_rep1} to generate a single representation, $h^1\in\mathbb{R}^{H\times C}$. 
This step allows the model to softly attend across all retrieved timesteps, promoting robustness to noise and preserving diverse temporal patterns.

While $h^1$ captures information from relevant past horizons, it may not align perfectly with the current forecast’s context. 
To retain fidelity to the base forecast $\widehat{x}^j$ while still benefiting from retrieval, we employ a second-stage fusion via a weighted skip connection:
\begin{align}
    \phi^1, \phi^2 &= \mathrm{softmax}\left(\tau\left(\widehat{x}^j\right)W^2, \tau\left(h^1\right)W^2\right)\label{eqn:weights2}\\
    h^2&=\phi^1\widehat{x}^j + \phi^2h^1\label{eqn:final_rep2}
\end{align}
where $\tau(\cdot)$ flattens its input, $\tau(\widehat{x}^j),\tau(h^1)\in\mathbb{R}^{HC},W^2\in\mathbb{R}^{HC\times 1}$ is a learnable parameter, $\phi^1,\phi^2\in\mathbb{R}$ are the weights and $h^2\in\mathbb{R}^{H\times C}$ is the final forecast. 
Eqn. \ref{eqn:final_rep2} provides an adaptive interpolation between the base forecast and the retrieved representation with the weights being trained under the anomaly prediction loss and anomaly upweighted forecasting loss. 
This ensures that our aggregation layer is tailored to fall back on the forecast if retrieved signals are noisy or less relevant with respect to anomaly prediction.

Our two-stage aggregation design is key to improving the forecast. 
The first stage ensures that retrieval is consolidated meaningfully, extracting generalizable anomaly cues. 
The second stage acts as a weighted skip connection, allowing the model to modulate how much it trusts the retrieval versus its own forecast. 
This structure enables flexibility and robustness, particularly important for anomaly prediction where retrieved horizons may vary in quality.

\begin{table*}[ht]
    \tiny
    \centering
    \caption{Performance comparison, in terms of VUS-PR $(\%)$, of multiple TSFM trained with F2A against AnomalyTransformer and no-RAG baselines. 
    Dataset abbreviations: CC - CreditCard, Dn - Daphnet, Gen - Genesis, OPP - Opportunity, and Exath - Exathlon. 
    Method abbreviations: Mom. - Moment, TSP. - TSPulse. 
    RAG$_k$ indicates the use of the top-k retrieved sequences in the RAG module, while RAG$_0$ denotes the F2A variant without retrieval. 
    The best values are highlighted in bold.}
    \label{tab:non_zero_benchmark}
    \addtolength{\tabcolsep}{-0.165em}
    \begin{tabular}{lllllllllllllllll}
    \toprule
         & \multicolumn{6}{c}{Zero-shot} \\
    \cmidrule{2-7}
        Method & \multicolumn{1}{c}{Gecco} & \multicolumn{1}{c}{PSM} & \multicolumn{1}{c}{Dn} & 
         \multicolumn{1}{c}{Gen} & \multicolumn{1}{c}{SWaT} & \multicolumn{1}{c}{CC} & \multicolumn{1}{c}{GHL}& \multicolumn{1}{c}{OPP}& \multicolumn{1}{c}{SMAP}& \multicolumn{1}{c}{MSL}& \multicolumn{1}{c}{MITDB}& \multicolumn{1}{c}{SVDB}& \multicolumn{1}{c}{Exath}& \multicolumn{1}{c}{SMD}& \multicolumn{1}{c}{LTDB}& \multicolumn{1}{c}{TAO}\\
    \toprule
    Mom.+AT & $03.27$ & $17.34$ & $\mathbf{09.60}$ & $01.45$ & $16.20$ & $05.00$ & $01.63$ & $02.34$ & $05.50$ & $10.39$ & $03.12$ & $07.06$ & $13.04$ & $05.37$ & $20.09$ & $83.87$ \\
    Mom.+RAG$_0$ & $03.81$ & $18.55$ & $05.55$ & $01.18$ & $\mathbf{32.87}$ & $06.16$ & $\mathbf{04.28}$ & $02.29$ & $\mathbf{29.93}$ & $14.81$ & $05.09$ & $\mathbf{35.21}$ & $95.57$ & $09.35$ & $40.64$ & $\mathbf{88.26}$ \\
    Mom.+RAG$_3$ & $04.30$ & $18.49$ & $05.73$ & $01.68$ & $22.95$ & $\mathbf{07.00}$ & $03.57$ & $02.54$ & $28.30$ & $\mathbf{15.34}$ & $05.07$ & $33.90$ & $\mathbf{96.88}$ & $09.24$ & $\mathbf{44.58}$ & $87.87$ \\
    Mom.+RAG$_5$ & $\mathbf{04.31}$ & $\mathbf{19.23}$ & $06.14$ & $\mathbf{01.77}$ & $23.97$ & $06.22$ & $03.60$ & $\mathbf{03.15}$ & $22.41$ & $14.50$ & $05.09$ & $33.87$ & $96.25$ & $09.44$ & $41.45$ & $87.47$ \\
    Mom.+RAG$_7$ & $03.77$ & $18.97$ & $07.22$ & $01.31$ & $28.78$ & $06.21$ & $03.45$ & $02.47$ & $26.66$ & $15.18$ & $\mathbf{05.15}$ & $33.85$ & $96.07$ & $\mathbf{09.51}$ & $41.25$ & $87.76$ \\
    \midrule
    TSP.+AT & $03.29$ & $17.42$ & $09.54$ & $01.40$ & $16.31$ & $05.00$ & $01.62$ & $02.33$ & $05.47$ & $10.32$ & $03.11$ & $07.03$ & $12.88$ & $05.35$ & $20.00$ & $83.82$ \\
    TSP.+RAG$_0$ & $04.98$ & $18.00$ & $07.25$ & $01.01$ & $\mathbf{37.09}$ & $06.20$ & $\mathbf{06.04}$ & $03.75$ & $24.09$ & $17.14$ & $05.93$ & $\mathbf{28.94}$ & $92.21$ & $07.31$ & $34.12$ & $87.67$ \\
    TSP.+RAG$_3$ & $\mathbf{06.92}$ & $\mathbf{18.96}$ & $09.88$ & $01.22$ & $32.09$ & $05.59$ & $04.15$ & $\mathbf{04.39}$ & $24.15$ & $\mathbf{19.03}$ & $05.56$ & $25.42$ & $92.38$ & $07.64$ & $39.39$ & $86.96$ \\
    TSP.+RAG$_5$ & $06.76$ & $18.62$ & $\mathbf{15.85}$ & $00.95$ & $29.49$ & $\mathbf{07.80}$ & $03.84$ & $03.53$ & $25.29$ & $16.73$ & $06.06$ & $27.75$ & $\mathbf{93.44}$ & $08.17$ & $\mathbf{39.96}$ & $85.59$ \\
    TSP.+RAG$_7$ & $05.74$ & $18.40$ & $09.30$ & $\mathbf{01.65}$ & $32.54$ & $07.30$ & $03.64$ & $02.59$ & $\mathbf{26.15}$ & $17.40$ & $\mathbf{06.09}$ & $25.34$ & $91.94$ & $\mathbf{08.48}$ & $36.01$ & $\mathbf{87.87}$ \\
    \midrule
    TTM+AT & $03.68$ & $05.29$ & $\mathbf{09.98}$ & $01.40$ & $18.47$ & $04.96$ & $03.68$ & $02.37$ & $05.45$ & $10.28$ & $03.59$ & $06.88$ & $12.00$ & $05.29$ & $20.88$ & $85.09$ \\
    TTM+RAG$_0$ & $05.84$ & $17.31$ & $05.26$ & $01.14$ & $68.77$ & $10.35$ & $\mathbf{04.86}$ & $04.75$ & $22.36$ & $19.3$ & $\mathbf{04.28}$ & $23.22$ & $92.79$ & $08.23$ & $33.76$ & $\mathbf{86.11}$ \\
    TTM+RAG$_3$ & $\mathbf{08.03}$ & $\mathbf{17.85}$ & $05.65$ & $05.55$ & $67.75$ & $\mathbf{11.10}$ & $04.41$ & $\mathbf{08.15}$ & $22.92$ & $17.96$ & $04.01$ & $\mathbf{25.79}$ & $92.03$ & $08.37$ & $\mathbf{40.05}$ & $85.48$ \\
    TTM+RAG$_5$ & $07.74$ & $17.67$ & $05.49$ & $04.68$ & $69.09$ & $08.75$ & $03.98$ & $06.72$ & $\mathbf{28.04}$ & $\mathbf{20.99}$ & $04.07$ & $25.42$ & $92.07$ & $\mathbf{08.93}$ & $38.14$ & $85.85$ \\
    TTM+RAG$_7$ & $07.34$ & $17.70$ & $05.32$ & $\mathbf{14.94}$ & $\mathbf{69.45}$ & $09.55$ & $04.07$ & $04.15$ & $25.34$ & $17.92$ & $04.13$ & $25.74$ & $\mathbf{93.01}$ & $08.60$ & $38.66$ & $85.71$ \\
    \bottomrule
    \end{tabular}
\end{table*}

\subsection{Forecast2Anomaly (F2A) Framework}
We now present the full architecture of our proposed Forecast2Anomaly (F2A) framework. 
Its overview is provided in Fig.~\ref{fig:f2a}.
We assume that $f_{\mathrm{f}}$ is a TSFM with an encoder-decoder architecture.
Given a sliding input window, $x\in\mathbb{R}^{L\times C_i}$ we first apply our channel selection strategy to choose a relevant subset of channels, followed by standard normalization. 
This yields a transformed signal $\widetilde{x}\in\mathbb{R}^{L\times C}$, which is passed to the encoder of the TSFM. 
The encoder is kept frozen throughout training, as it captures generalized temporal representations learned from large-scale pretraining corpora. 
Only the decoder and downstream components are fine-tuned for the anomaly prediction task. 
The encoder output is then fed into the decoder of $f_{\mathrm{f}}$, followed by a projection layer that generates forecasts over a fixed horizon. 
Let $\widehat{x} \in \mathbb{R}^{H \times C}$ denote this forecast.

To augment the forecast with relevant contextual priors, we retrieve $k$ historical horizons from a retrieval database using the embedding of $\widetilde{x}$ generated from the encoder.
Let the retrieved set be denoted as $\{z^q\}_{q=1}^k$, where each $z^q \in \mathbb{R}^{H \times C}$.
However, direct fusion of $\widehat{x}$ and ${z^q}$ can be suboptimal due to possible scale mismatch between the TSFM forecast and the retrieved sequences. 
To address this, we develop a \textbf{scaling layer} which applies a learnable scaling transformation to $\widehat{x}$ before fusion. 
Specifically, we learn a globally-aware representation by flattening $\widehat{x}$ into a vector, transforming it with a fully connected layer, and reshaping it back, in the following manner,
\begin{equation}
\widehat{x}^s = \tau^{-1}\left(\tau\left(\widehat{x}\right)W^s\right)
\end{equation}
Here, $\tau(\cdot)$ flattens the forecast $\widehat{x} \in \mathbb{R}^{H \times C}$ into a vector of dimension $HC$, and $\tau^{-1}(\cdot)$ reshapes the result back to the original $H \times C$ form. 
The learnable matrix $W^s \in \mathbb{R}^{HC \times HC}$ enables the model to capture global inter-channel and inter-temporal dependencies during scaling.
Note that this is not a normalization step, but rather a transformation that learns a new forecast representation more compatible with the retrieved horizons.

We then fuse the scaled forecast $\widehat{x}^s$ with the retrieved set ${z^q}$ using an aggregation module, resulting in an enriched forecast $\widehat{x}^f \in \mathbb{R}^{H \times C}$. 
The fusion mechanism allows the model to refine its predictions by incorporating complementary temporal patterns from the retrieval bank.
Finally, the refined forecast $\widehat{x}^f$ is passed through the \textbf{anomaly prediction layer} to produce per-timestep anomaly probabilities in the following manner,
\begin{equation}
    p = \sigma\left(\tau\left(\widehat{x}^f\right)W^{\mathrm{ap}}\right)
\end{equation}
where $p\in\mathbb{R}^{H}$ is the vector of anomaly probabilities for each timestep, $\sigma(\cdot)$ is a sigmoid function, $W^{\mathrm{ap}}\in\mathbb{R}^{HC\times H}$ is a learnable parameter and $\tau(\cdot)$ is the flattening function. 

During inference, the model directly outputs $p$ given the input window $x$. 
During training, however, the model additionally receives the true future horizon $z$ and the anomaly label sequence $y$, and is trained to minimize the joint loss $\mathcal{L}$ over both forecasting and anomaly prediction objectives.

\begin{table*}[ht]
    \tiny
    \centering
    \caption{Performance comparison, in terms of VUS-PR $(\%)$, of the best F2A trained TSFM, against statistical methods.
    Dataset abbreviations: CC - CreditCard, Dn - Daphnet, Gen - Genesis. 
    F2A$_{\text{ours}}$ denote the best-performing configuration of our framework across all TSFMs.  
    The highest values are highlighted in bold.}
    \label{tab:statistical_benchmark}
    \addtolength{\tabcolsep}{-0.15em}
    \begin{tabular}{lllllllllllllllll}
    \toprule
         & \multicolumn{6}{c}{Zero-shot} \\
    \cmidrule{2-7}
        Method & \multicolumn{1}{c}{Gecco} & \multicolumn{1}{c}{PSM} & \multicolumn{1}{c}{Dn} & 
         \multicolumn{1}{c}{Gen} & \multicolumn{1}{c}{SWaT} & \multicolumn{1}{c}{CC}& \multicolumn{1}{c}{GHL}& \multicolumn{1}{c}{OPP}& \multicolumn{1}{c}{SMAP}& \multicolumn{1}{c}{MSL}& \multicolumn{1}{c}{MITDB}& \multicolumn{1}{c}{SVDB}& \multicolumn{1}{c}{Exath}& \multicolumn{1}{c}{SMD}& \multicolumn{1}{c}{LTDB}& \multicolumn{1}{c}{TAO}\\
    \toprule
    IForest & $03.05$ & $18.86$ & $16.40$ & $00.52$ & $11.17$ & $07.89$ & $01.58$ & $03.12$ & $02.76$ & $12.38$ & $03.12$ & $15.19$ & $10.98$ & $08.15$ & $22.76$ & $87.27$ \\
    CBLOF & $03.09$ & $17.83$ & $\mathbf{35.15}$ & $05.61$ & $11.30$ & $05.41$ & $02.30$ & $02.27$ & $02.97$ & $14.23$ & $03.45$ & $09.18$ & $30.80$ & $04.74$ & $24.81$ & $\mathbf{99.99}$\\
    RobustPCA & $03.71$ & $19.09$ & $08.34$ & $05.77$ & $10.79$ & $06.63$ & $00.91$ & $02.34$ & $03.33$ & $05.52$ & $03.09$ & $08.16$ & $28.37$ & $05.04$ & $19.36$ & $81.68$\\
    KMeansAD & $\mathbf{08.44}$ & $16.65$ & $22.76$ & $00.56$ & $13.08$ & $05.47$ & $03.00$ & $04.31$ & $03.15$ & $08.40$ & $03.72$ & $06.88$ & $37.00$ & $04.50$ & $19.47$ & $94.49$\\
    F2A$_{\text{ours}}$ & $08.03$  & $\mathbf{19.23}$ & $15.85$ & $\mathbf{14.94}$ & $\mathbf{69.45}$ & $\mathbf{11.10}$ & $\mathbf{06.04}$ & $\mathbf{08.15}$ & $\mathbf{29.93}$ & $\mathbf{20.99}$ & $\mathbf{06.09}$ & $\mathbf{35.21}$ & $\mathbf{96.88}$ & $\mathbf{09.51}$ & $\mathbf{44.58}$ & $88.28$\\
    \bottomrule
    \end{tabular}
\end{table*}

\section{Experimental Setup}
\subsection{Dataset and RAG Database}
We conduct experiments on $16$ diverse multivariate time series datasets drawn from the recently introduced TSB-AD-M benchmark~\cite{liu2024elephant}, which offers a comprehensive evaluation suite for anomaly detection in time series. 
We use the official train split for fine-tuning as well as populating the RAG database. 
For broad coverage of the RAG database, we also consider the first $30\%$ of samples in the official test split, per dataset. 
To avoid data leakage, all evaluations are performed on the remaining $70\%$. 
The embeddings for the input windows are generated using the TSPulse encoder for all experiments.

A subset of $6$ datasets, namely, GECCO, PSM, Genesis, Daphnet, SWaT, and CreditCard are officially excluded from the train set by the TSB-AD-M benchmark and are used exclusively for evaluation. 
Hence, these datasets serve as zero-shot testbeds, allowing us to gauge the generalization ability of our approach.

\subsection{Baselines and Evaluation Metrics}
Our proposed framework is designed to be adaptable to any encoder-decoder-based TSFM for anomaly prediction. To evaluate We select three representative TSFMs as forecasting backbones ($f_{\mathrm{f}}$), namely, TTM (TinyTimeMixer) \cite{ekambaram2024tiny}, TSPulse \cite{ekambaram2025tspulse}, and Moment \cite{goswami2024moment}. 
In our framework, the anomaly prediction module ($f_{\mathrm{ap}}$) is implemented via a learned parameter matrix $W^{\mathrm{ap}}$ that transforms the forecasting error into anomaly predictions.
For a fair comparison, we construct baseline models by retaining the same three TSFMs as forecasters but replacing our anomaly prediction head with AnomalyTransformer \cite{xu2021anomaly}, a widely-used reconstruction-based anomaly detection model. 
We also provide an additional baseline which is our framework but without the RAG module.
For a comprehensive comparison we benchmark our best results against well-known, competitive statistical methods, namely IsolationForest \cite{liu2008isolation}, CBLOF \cite{he2003discovering}, RobustPCA \cite{paffenroth2018robust} and KMeansAD \cite{paparrizos2017fast}. 
Thus, we compare F2A against $10$ separate baselines.

We follow the official evaluation protocol of the TSB-AD-M benchmark \cite{liu2024elephant}, utilizing its metrics computation library to ensure consistency and reproducibility. 
Our primary evaluation metric is the Volume Under the Surface for Precision-Recall (VUS-PR) \cite{paparrizos2022volume}, which the benchmark highlights as a robust metric to thresholding biases. 
We also compute standard F1, Precision, and Recall scores, which are reported in Section 1 of the Supplementary Materials. 
For reproducibility, all hyperparameters and implementation details are provided in Section 2 of the Supplementary Materials.

\section{Results}
\subsection{Performance on Zero-Shot Benchmarks}
To assess F2A's generalization, we evaluate it in a zero-shot setting on datasets entirely excluded from fine-tuning, namely, GECCO, PSM, Daphnet, Genesis, SWaT, and CreditCard.
As shown in Table \ref{tab:non_zero_benchmark}, F2A consistently outperforms the strong AnomalyTransformer (AT) baseline across five of the six benchmarks with the Mometum model.
When compared against the non-RAG variant, our retrieval-augmented generation (RAG) approach yields consistent improvements across all six datasets, demonstrating the benefit of leveraging similar historical horizons.
With the TSPulse backbone, F2A outperforms AT on all six datasets, and the RAG-enhanced variant beats its non-RAG counterpart on five out of six. 
For the compact TinyTimeMixer (TTM) model, F2A surpasses AT on five datasets, and its retrieval-augmented version outperforms the baseline on all six, highlighting the effectiveness of our method even in low-capacity regimes.

% \textbf{F2A with Moment:}
% F2A achieves a mean absolute gain of $+2.6\%$ over AT, with a peak improvement of $+12.5\%$ on SWaT. 
% It outperforms AT on $5$ out of $6$ datasets, and improves over its non-RAG version (RAG$_0$) in all but one case. 
% Even F2A (RAG$_0$) outperforms AT by $+16.7\%$ on SWaT.

% \textbf{F2A with TSPulse:}
% F2A delivers a $+5.13\%$ average gain over AT, outperforming it on all $6$ datasets. 
% The largest improvements are on Daphnet $(+6.31\%)$ and SWaT $(+16.23\%)$. 
% Retrieval enhances performance in $5$ out of $6$ datasets, with a maximum RAG boost of $+8.6\%$ on Daphnet.

% \textbf{F2A with TinyTimeMixer (TTM):}
% Despite TTM’s compact architecture, F2A attains a remarkable $+13.87\%$ average improvement over AT, outperforming it on $5$ out of $6$ datasets. 
% SWaT sees a dramatic gain of $+50.98\%$, while retrieval boosts Genesis by $+13.8\%$, improving performance in all datasets.

% In summary, F2A excels in zero-shot anomaly prediction, where RAG-driven refinement significantly enhances forecasting across diverse and unseen domains, especially for lightweight forecasters like TTM.

\subsection{Performance on Non-Zero-Shot Benchmarks}
To assess the benefits of F2A in data-rich scenarios, we evaluate its performance in a non-zero-shot setting, where models are fine-tuned on each target dataset. 
This allows us to examine whether RAG-based augmentation and model adaptation provide additional gains even when domain-specific training is allowed. 
As reported in Table~\ref{tab:non_zero_benchmark}, F2A consistently improves forecasting accuracy across multiple datasets and forecasters.
Specifically, the Momentum model when adapted to the task of anomaly prediction using our framework outperforms the AT baseline across all ten datasets. 
Against the non-RAG version, the full F2A setup (with RAG) outperforms it over six different datasets. 
With TSPulse model, F2A outperforms AT across all ten datasets while achieving better performance over eight datasets against the non-RAG variant. 
Similarly, F2A outperforms AT across all ten datasets. 
On the other hand, the entire F2A setup (including RAG) outperforms the non-RAG variant over seven different datasets.

% \textbf{F2A with Moment:}
% F2A yields a substantial average improvement of $+17.58\%$ over the anomaly thresholding (AT) baseline, achieving a striking gain of $+83.84\%$ on Exathlon. 
% It surpasses AT on all $10$ datasets and improves upon its non-retrieval counterpart (RAG$_0$) in $7$ cases, with the largest RAG-based boost observed on LTDB ($+3.94\%$). 
% Notably, even without retrieval, F2A (RAG$_0$) exceeds AT by $+82.53\%$ on Exathlon.

% \textbf{F2A with TSPulse:}
% When paired with TSPulse, F2A achieves an average gain of $+16.54\%$ over AT, outperforming it on all $10$ datasets. 
% The most significant gains occur on Exathlon ($+80.56\%$), SMAP ($+20.68\%$), and SVDB ($+20.72\%$). Retrieval enhances performance on $8$ datasets against its non-retrieval counterpart, with the highest gain again on LTDB ($+5.84\%$).

% \textbf{F2A with TinyTimeMixer (TTM):}
% F2A achieves an average improvement of $+16.35\%$ over AT, again outperforming it across all $10$ benchmarks. 
% Exathlon and LTDB record the most pronounced improvements at $+81.01\%$ and $+19.17\%$, respectively. 
% Retrieval further boosts performance on $7$ datasets, with LTDB seeing the highest gain of $+6.29\%$.

In summary, F2A demonstrates strong performance in non-zero-shot scenarios as well, delivering consistent gains over the AT baseline. 
Its benefits persist across forecasters of varying capacity, suggesting that RAG-based augmentation and fusion not only help generalization but also enhance learning even when training data is available.

\subsection{Performance against Statistical Methods}
To benchmark F2A’s zero-shot anomaly prediction against traditional unsupervised detectors, we compare our best F2A configuration (across all TSFMs) to four widely used statistical methods, namely, Isolation Forest (IForest), Cluster-Based Local Outlier Factor (CBLOF), Robust PCA (RPCA), and KMeansAD (KMAD). Table~\ref{tab:statistical_benchmark} reports the comparison results.
Despite the simplicity of classical methods, KMAD achieves the highest score on GECCO $(8.44\%)$ and CBLOF on Daphnet $(35.15\%)$ and TAO $(99.99\%)$. 
However, F2A$_{\text{ours}}$ delivers top performance on $13$ of $16$ datasets, including an absolute improvement of $+58.88\%$ on the Exathlon dataset against the best statistical baseline.

This comparison highlights the fact that traditional detectors struggle with high-dimensional, evolving time series in a zero-shot context while F2A’s integration of TSFM forecasting with adaptive retrieval yields substantial gains on nearly all datasets, establishing a new state-of-the-art for zero-shot anomaly prediction.

\begin{figure*}[t]
    \centering
    \begin{subfigure}{0.31\textwidth}
        \centering
        \includegraphics[width=\linewidth]{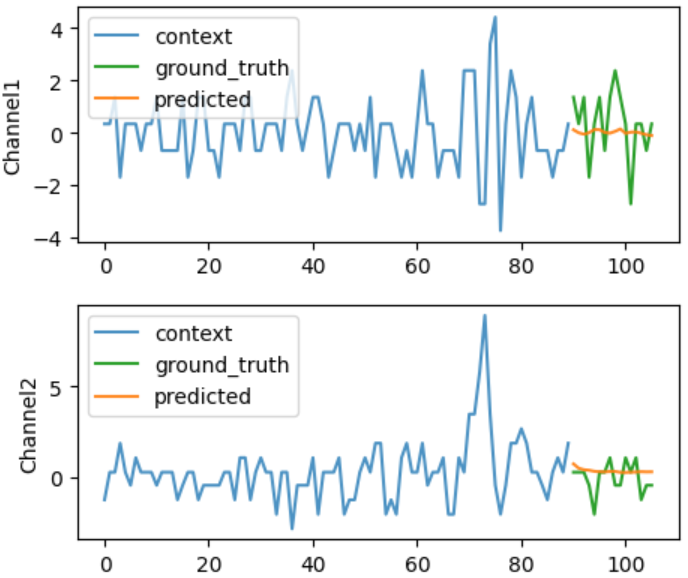}
        \caption{Vanilla (no fine-tuning) TSPulse Forecast}
        \label{fig:daphnet_vanilla_forecast}
    \end{subfigure}
    \hfill
    \begin{subfigure}{0.31\textwidth}
        \centering
        \includegraphics[width=\linewidth]{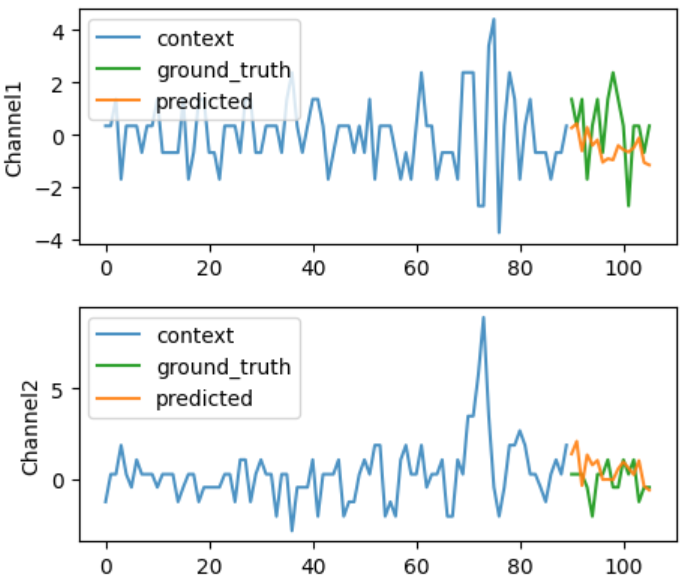}
        \caption{TSPulse fine-tuned using F2A without RAG}
        \label{fig:daphnet_norag_forecast}
    \end{subfigure}
    \hfill
    \begin{subfigure}{0.31\textwidth}
        \centering
        \includegraphics[width=\linewidth]{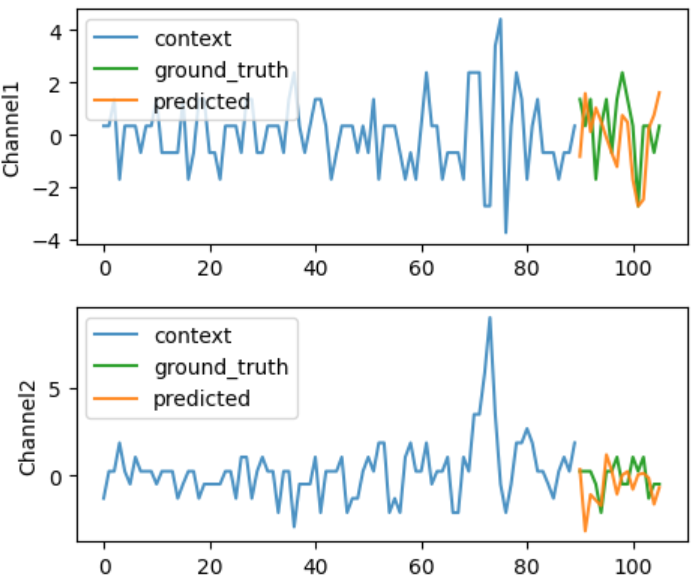}
        \caption{TSPulse fine-tuned using full F2A setup (with RAG)}
        \label{fig:daphnet_rag_forecast}
    \end{subfigure}
    \caption{Forecast comparison on a randomly picked example from the Daphnet dataset on three channels. Forecasts from three versions of TSPulse: (a) vanilla (no fine-tuning), (b) fine-tuned using F2A without RAG, and (c) fine-tuned using F2A with RAG. 
    Green denotes the ground truth forecast, orange denotes the predicted output, and blue is the context window.}
    \label{fig:forecast_comparison}
\end{figure*}

\section{Discussion}
\subsection{Forecast Loss Improves Anomaly Prediction}
To assess the contribution of the forecasting loss $\mathcal{L}_f$, we conduct a targeted ablation by evaluating F2A with TSPulse as the base forecaster on zero-shot benchmarks, once with $\lambda=0$ (no forecasting loss) and once with $\lambda=1$. 
The results, summarized in Table~\ref{tab:loss_ablation_no_forecast}, clearly demonstrate that removing $\mathcal{L}_f$ leads to a substantial degradation in anomaly detection performance across nearly all datasets. 
%establishing the crucial role that forecasting loss plays in preserving anomaly-specific temporal signatures. 
In summary, this ablation establishes that the forecasting loss plays a crucial role in preserving anomaly-specific temporal signatures. 
By aligning the learned latent representations with both forecasting and anomaly objectives, $\mathcal{L}_f$ prevents over-smoothing and promotes anomaly-aware representations.

\begin{table}[H]
    % \tiny
    \centering
    \caption{Impact of the \textit{forecasting loss} on zero-shot anomaly prediction using TSPulse, during training, by comparing F2A with $\lambda=0$ and $\lambda=1$.  
    F2A$_{\lambda=0}$ means training without $\mathcal{L}_f$. 
    All results are percentages in terms of VUS-PR.}
    \label{tab:loss_ablation_no_forecast}
    \addtolength{\tabcolsep}{-0.25em}
    \begin{tabular}{lllllll}
    \toprule
    Method & \multicolumn{1}{c}{Gecco} & \multicolumn{1}{c}{PSM} & \multicolumn{1}{c}{Dn} & 
     \multicolumn{1}{c}{Gen} & \multicolumn{1}{c}{SWaT} & \multicolumn{1}{c}{CC} \\
    \toprule
    F2A$_{\lambda=0}$ & $04.77$ & $18.88$ & $05.38$ & $00.48$ & $28.34$ & $4.92$\\
    F2A$_{\lambda=1}$ & $\mathbf{06.92}$ & $\mathbf{18.96}$ & $\mathbf{09.88}$ & $\mathbf{01.22}$ & $\mathbf{32.09}$ & $\mathbf{05.59}$\\
    \bottomrule
    \end{tabular}
\end{table}

\subsection{Weighted Forecast Loss Improves Anomaly Expressiveness}
To evaluate the impact of emphasizing forecasting errors for anomalous time steps, we perform a targeted ablation on the weighting mechanism in the forecasting loss $\mathcal{L}_f$. 
Specifically, we compare two variants of F2A using TSPulse on zero-shot benchmarks, one where the loss treats all time steps equally ($\psi = 1$), and another where forecasting errors during anomalies are upweighted ($\psi > 1$).
As shown in Table~\ref{tab:loss_ablation_no_weight_forecast}, assigning greater importance to anomalous regions consistently increases VUS-PR scores in almost all data sets. 
%In essence, 
Our weighting strategy 
%This 
enhances the expressiveness of anomalies by preventing their dilution in the forecast.%, which is crucial for downstream identification tasks.

\begin{table}[H]
    % \tiny
    \centering
    \caption{Impact of the weights in forecasting loss on zero-shot anomaly prediction, during training, by comparing F2A with $\psi=1$ and $\psi\geq1$.  
    F2A$_{]\psi=1}$ means training without giving more weight to anomalies in the forecast.
    All results are percentages in terms of VUS-PR.} 
    %The best values for each setting are highlighted in bold.}
    \label{tab:loss_ablation_no_weight_forecast}
    \addtolength{\tabcolsep}{-0.22em}
    \begin{tabular}{lllllll}
    \toprule
    Method & \multicolumn{1}{c}{Gecco} & \multicolumn{1}{c}{PSM} & \multicolumn{1}{c}{Dn} & 
     \multicolumn{1}{c}{Gen} & \multicolumn{1}{c}{SWaT} & \multicolumn{1}{c}{CC} \\
    \toprule
    F2A$_{\psi=1}$ & $05.78$ & $\mathbf{19.01}$ & $05.52$ & $00.52$ & $26.62$ & $04.25$\\
    F2A$_{\psi=3}$ & $\mathbf{06.92}$ & $18.96$ & $\mathbf{09.88}$ & $\mathbf{01.22}$ & $\mathbf{32.09}$ & $\mathbf{05.59}$\\
    \bottomrule
    \end{tabular}
\end{table}

\subsection{F2A Removes Anomaly Suppression in TSFM Forecast}
To understand the source of performance gains observed with our RAG module, particularly in the zero-shot setting (Table \ref{tab:non_zero_benchmark}), we analyze the forecasts of TSPulse under different configurations on the Daphnet dataset. Specifically, we compare:
(a) a vanilla TSPulse model (no fine-tuning), (b) TSPulse fine-tuned using F2A without RAG, and (c) TSPulse fine-tuned using F2A with RAG.

As shown in Fig.~\ref{fig:forecast_comparison}, there is a clear progression in forecast accuracy across the three setups. 
The vanilla model fails to align with the ground truth in the anomalous region demonstrating the over-smoothing problem. 
On the other hand, F2A fine-tuning improves forecast alignment while, most notably, the full F2A setup with RAG exhibits tight correspondence between predicted and ground truth signals in the forecast horizon, especially around the anomalous segment. 
This qualitative evidence supports our hypothesis that improved anomaly prediction is tightly coupled with improvement in forecasting.

%\section{Conclusion}
\section{Conclusion and Future Work}
Forecast2Anomaly (F2A) advances multivariate time series analysis by adapting pre-trained Time Series Foundation Models for proactive anomaly prediction through a joint forecast-anomaly loss that preserves irregular patterns and an anomaly sensitive Retrieval-Augmented Generation (RAG) module that dynamically incorporates relevant historical contexts without retraining. Extensive evaluations on $16$ diverse datasets including $6$ zero-shot benchmarks and three TSFM backbones demonstrate that F2A significantly outperforms state-of-the-art baselines in both zero-shot and fine-tuned settings, establishing a new standard for scalable, generalizable early warning systems.

Despite these gains, F2A’s efficacy depends on the breadth and representativeness of its retrieval database and introduces inference overhead that may challenge low-latency applications. Its fixed-window forecasting may miss anomalies occurring at varied time scales, and channel selection via variance could overlook subtle signals. Future work will explore distinguishing between known versus novel anomalies to better assess model confidence on unseen failure patterns, adaptive or multi-scale prediction horizons, more sophisticated retrieval strategies (e.g., causal or semantic similarity), and online updating mechanisms to continuously incorporate emerging anomaly behaviors.

%\section{Acknowledgments}
% Write acknowledgements here.

\bigskip

\bibliography{aaai2026}

\begin{thebibliography}{29}
\providecommand{\natexlab}[1]{#1}

\bibitem[{Ansari et~al.(2024)Ansari, Stella, T{\"{u}}rkmen, Zhang, Mercado, Shen, Shchur, Rangapuram, Pineda{-}Arango, Kapoor, Zschiegner, Maddix, Wang, Mahoney, Torkkola, Wilson, Bohlke{-}Schneider, and Wang}]{ansarichronos}
Ansari, A.~F.; Stella, L.; T{\"{u}}rkmen, A.~C.; Zhang, X.; Mercado, P.; Shen, H.; Shchur, O.; Rangapuram, S.~S.; Pineda{-}Arango, S.; Kapoor, S.; Zschiegner, J.; Maddix, D.~C.; Wang, H.; Mahoney, M.~W.; Torkkola, K.; Wilson, A.~G.; Bohlke{-}Schneider, M.; and Wang, B. 2024.
\newblock Chronos: Learning the Language of Time Series.
\newblock \emph{Trans. Mach. Learn. Res.}, 2024.

\bibitem[{Das et~al.(2024)Das, Kong, Sen, and Zhou}]{das2024decoder}
Das, A.; Kong, W.; Sen, R.; and Zhou, Y. 2024.
\newblock A decoder-only foundation model for time-series forecasting.
\newblock In \emph{Forty-first International Conference on Machine Learning}.

\bibitem[{Ekambaram et~al.(2024)Ekambaram, Jati, Dayama, Mukherjee, Nguyen, Gifford, Reddy, and Kalagnanam}]{ekambaram2024tiny}
Ekambaram, V.; Jati, A.; Dayama, P.; Mukherjee, S.; Nguyen, N.; Gifford, W.~M.; Reddy, C.; and Kalagnanam, J. 2024.
\newblock Tiny time mixers (ttms): Fast pre-trained models for enhanced zero/few-shot forecasting of multivariate time series.
\newblock \emph{Advances in Neural Information Processing Systems}, 37: 74147--74181.

\bibitem[{Ekambaram et~al.(2025)Ekambaram, Kumar, Jati, Mukherjee, Sakai, Dayama, Gifford, and Kalagnanam}]{ekambaram2025tspulse}
Ekambaram, V.; Kumar, S.; Jati, A.; Mukherjee, S.; Sakai, T.; Dayama, P.; Gifford, W.~M.; and Kalagnanam, J. 2025.
\newblock TSPulse: Dual Space Tiny Pre-Trained Models for Rapid Time-Series Analysis.
\newblock \emph{arXiv preprint arXiv:2505.13033}.

\bibitem[{Garza, Challu, and Mergenthaler-Canseco(2023)}]{garza2023timegpt}
Garza, A.; Challu, C.; and Mergenthaler-Canseco, M. 2023.
\newblock TimeGPT-1.
\newblock \emph{arXiv preprint arXiv:2310.03589}.

\bibitem[{Goswami et~al.(2024)Goswami, Szafer, Choudhry, Cai, Li, and Dubrawski}]{goswami2024moment}
Goswami, M.; Szafer, K.; Choudhry, A.; Cai, Y.; Li, S.; and Dubrawski, A. 2024.
\newblock Moment: A family of open time-series foundation models.
\newblock \emph{arXiv preprint arXiv:2402.03885}.

\bibitem[{Han et~al.(2025{\natexlab{a}})Han, Lee, Cha, Arik, and Yoon}]{hanretrieval}
Han, S.; Lee, S.; Cha, M.; Arik, S.~{\"{O}}.; and Yoon, J. 2025{\natexlab{a}}.
\newblock Retrieval Augmented Time Series Forecasting.
\newblock \emph{CoRR}, abs/2505.04163.

\bibitem[{Han et~al.(2025{\natexlab{b}})Han, Lee, Cha, Arik, and Yoon}]{han2025retrieval}
Han, S.; Lee, S.; Cha, M.; Arik, S.~O.; and Yoon, J. 2025{\natexlab{b}}.
\newblock Retrieval augmented time series forecasting.
\newblock \emph{arXiv preprint arXiv:2505.04163}.

\bibitem[{He, Xu, and Deng(2003)}]{he2003discovering}
He, Z.; Xu, X.; and Deng, S. 2003.
\newblock Discovering cluster-based local outliers.
\newblock \emph{Pattern recognition letters}, 24(9-10): 1641--1650.

\bibitem[{Lewis et~al.(2020)Lewis, Perez, Piktus, Petroni, Karpukhin, Goyal, K{\"u}ttler, Lewis, Yih, Rockt{\"a}schel et~al.}]{lewis2020retrieval}
Lewis, P.; Perez, E.; Piktus, A.; Petroni, F.; Karpukhin, V.; Goyal, N.; K{\"u}ttler, H.; Lewis, M.; Yih, W.-t.; Rockt{\"a}schel, T.; et~al. 2020.
\newblock Retrieval-augmented generation for knowledge-intensive nlp tasks.
\newblock \emph{Advances in neural information processing systems}, 33: 9459--9474.

\bibitem[{Liang et~al.(2024)Liang, Wen, Nie, Jiang, Jin, Song, Pan, and Wen}]{liang2024foundation}
Liang, Y.; Wen, H.; Nie, Y.; Jiang, Y.; Jin, M.; Song, D.; Pan, S.; and Wen, Q. 2024.
\newblock Foundation models for time series analysis: A tutorial and survey.
\newblock In \emph{Proceedings of the 30th ACM SIGKDD conference on knowledge discovery and data mining}, 6555--6565.

\bibitem[{Lin et~al.(2017)Lin, Goyal, Girshick, He, and Doll{\'a}r}]{lin2017focal}
Lin, T.-Y.; Goyal, P.; Girshick, R.; He, K.; and Doll{\'a}r, P. 2017.
\newblock Focal loss for dense object detection.
\newblock In \emph{Proceedings of the IEEE international conference on computer vision}, 2980--2988.

\bibitem[{Liu, Ting, and Zhou(2008)}]{liu2008isolation}
Liu, F.~T.; Ting, K.~M.; and Zhou, Z.-H. 2008.
\newblock Isolation forest.
\newblock In \emph{2008 eighth ieee international conference on data mining}, 413--422. IEEE.

\bibitem[{Liu et~al.(2024)Liu, Yang, Li, and Hong}]{liu2024retrieval}
Liu, J.; Yang, L.; Li, H.; and Hong, S. 2024.
\newblock Retrieval-augmented diffusion models for time series forecasting.
\newblock \emph{Advances in Neural Information Processing Systems}, 37: 2766--2786.

\bibitem[{Liu and Paparrizos(2024)}]{liu2024elephant}
Liu, Q.; and Paparrizos, J. 2024.
\newblock The elephant in the room: Towards a reliable time-series anomaly detection benchmark.
\newblock \emph{Advances in Neural Information Processing Systems}, 37: 108231--108261.

\bibitem[{Malhotra et~al.(2015)Malhotra, Vig, Shroff, and Agarwal}]{malhotra2015long}
Malhotra, P.; Vig, L.; Shroff, G.; and Agarwal, P. 2015.
\newblock Long Short Term Memory Networks for Anomaly Detection in Time Series.
\newblock In \emph{23rd European Symposium on Artificial Neural Networks, {ESANN} 2015, Bruges, Belgium, April 22-24, 2015}.

\bibitem[{Ning et~al.(2025)Ning, Pan, Liu, Jiang, Zhang, Rasul, Schneider, Ma, Nevmyvaka, and Song}]{ning2025ts}
Ning, K.; Pan, Z.; Liu, Y.; Jiang, Y.; Zhang, J.~Y.; Rasul, K.; Schneider, A.; Ma, L.; Nevmyvaka, Y.; and Song, D. 2025.
\newblock Ts-rag: Retrieval-augmented generation based time series foundation models are stronger zero-shot forecaster.
\newblock \emph{arXiv preprint arXiv:2503.07649}.

\bibitem[{Paffenroth, Kay, and Servi(2018)}]{paffenroth2018robust}
Paffenroth, R.; Kay, K.; and Servi, L. 2018.
\newblock Robust pca for anomaly detection in cyber networks.
\newblock \emph{arXiv preprint arXiv:1801.01571}.

\bibitem[{Paparrizos et~al.(2022)Paparrizos, Boniol, Palpanas, Tsay, Elmore, and Franklin}]{paparrizos2022volume}
Paparrizos, J.; Boniol, P.; Palpanas, T.; Tsay, R.~S.; Elmore, A.; and Franklin, M.~J. 2022.
\newblock Volume under the surface: a new accuracy evaluation measure for time-series anomaly detection.
\newblock \emph{Proceedings of the VLDB Endowment}, 15(11): 2774--2787.

\bibitem[{Paparrizos and Gravano(2017)}]{paparrizos2017fast}
Paparrizos, J.; and Gravano, L. 2017.
\newblock Fast and accurate time-series clustering.
\newblock \emph{ACM Transactions on Database Systems (TODS)}, 42(2): 1--49.

\bibitem[{Park et~al.(2025)Park, Lee, Kim, and Park}]{parkwill}
Park, M.; Lee, W.; Kim, S.~T.; and Park, G. 2025.
\newblock When Will It Fail?: Anomaly to Prompt for Forecasting Future Anomalies in Time Series.
\newblock \emph{CoRR}, abs/2506.23596.

\bibitem[{Sakurada and Yairi(2014)}]{sakurada2014anomaly}
Sakurada, M.; and Yairi, T. 2014.
\newblock Anomaly detection using autoencoders with nonlinear dimensionality reduction.
\newblock In \emph{Proceedings of the MLSDA 2014 2nd workshop on machine learning for sensory data analysis}, 4--11.

\bibitem[{Shyalika et~al.(2024)Shyalika, Bagga, Bhatt, Prasad, Ghazo, and Sheth}]{shyalika2024time}
Shyalika, C.; Bagga, H.~K.; Bhatt, A.; Prasad, R.; Ghazo, A.~A.; and Sheth, A. 2024.
\newblock Time series foundational models: Their role in anomaly detection and prediction.
\newblock \emph{arXiv preprint arXiv:2412.19286}.

\bibitem[{Woo et~al.(2024)Woo, Liu, Kumar, Xiong, Savarese, and Sahoo}]{woo2024unified}
Woo, G.; Liu, C.; Kumar, A.; Xiong, C.; Savarese, S.; and Sahoo, D. 2024.
\newblock Unified Training of Universal Time Series Forecasting Transformers.
\newblock In \emph{Forty-first International Conference on Machine Learning, {ICML} 2024, Vienna, Austria, July 21-27, 2024}. OpenReview.net.

\bibitem[{Xu et~al.(2021)Xu, Wu, Wang, and Long}]{xu2021anomaly}
Xu, J.; Wu, H.; Wang, J.; and Long, M. 2021.
\newblock Anomaly transformer: Time series anomaly detection with association discrepancy.
\newblock \emph{arXiv preprint arXiv:2110.02642}.

\bibitem[{Yang et~al.(2025)Yang, Wang, Zheng, and Jin}]{yang2025timerag}
Yang, S.; Wang, D.; Zheng, H.; and Jin, R. 2025.
\newblock Timerag: Boosting llm time series forecasting via retrieval-augmented generation.
\newblock In \emph{ICASSP 2025-2025 IEEE International Conference on Acoustics, Speech and Signal Processing (ICASSP)}, 1--5. IEEE.

\bibitem[{Yeh et~al.(2023)Yeh, Dai, Chen, Zheng, Fan, Der, Lai, Zhuang, Wang, Wang et~al.}]{yeh2023toward}
Yeh, C.-C.~M.; Dai, X.; Chen, H.; Zheng, Y.; Fan, Y.; Der, A.; Lai, V.; Zhuang, Z.; Wang, J.; Wang, L.; et~al. 2023.
\newblock Toward a foundation model for time series data.
\newblock In \emph{Proceedings of the 32nd ACM International Conference on Information and Knowledge Management}, 4400--4404.

\bibitem[{Zamanzadeh~Darban et~al.(2024)Zamanzadeh~Darban, Webb, Pan, Aggarwal, and Salehi}]{zamanzadeh2024deep}
Zamanzadeh~Darban, Z.; Webb, G.~I.; Pan, S.; Aggarwal, C.; and Salehi, M. 2024.
\newblock Deep learning for time series anomaly detection: A survey.
\newblock \emph{ACM Computing Surveys}, 57(1): 1--42.

\bibitem[{Zhong et~al.(2023)Zhong, Fan, Zhang, Ma, Zhang, Sun, Lin, Zhang, and Pei}]{zhong2023survey}
Zhong, Z.; Fan, Q.; Zhang, J.; Ma, M.; Zhang, S.; Sun, Y.; Lin, Q.; Zhang, Y.; and Pei, D. 2023.
\newblock A survey of time series anomaly detection methods in the aiops domain.
\newblock \emph{arXiv preprint arXiv:2308.00393}.

\end{thebibliography}

% \end{document}

\clearpage
\appendix
\clearpage
\onecolumn % switch to one column for the supplement title
\begin{center}
\LARGE \textbf{Supplementary Material}

    %\LARGE \textbf{Supplementary Material for Forecast2Anomaly (F2A):
    %Adapting Multivariate Time Series Foundation Models for Anomaly Prediction}
\end{center}
\vspace{1em}

\section*{F1-score supplementing Table 1}
\begin{table*}[h]
    \tiny
    \centering
    \caption{Performance comparison, in terms of F1-score $(\%)$, of multiple TSFM trained with F2A across various topk values of RAG. 
    Dataset abbreviations: CC - CreditCard, Dn - Daphnet, Gen - Genesis, OPP - Opportunity, and Exath - Exathlon. 
    Method abbreviations: Mom. - Moment, TSP. - TSPulse. 
    RAG$_k$ indicates the use of the top-k retrieved sequences in the RAG module, while RAG$_0$ denotes the F2A variant without retrieval. 
    The best values are highlighted in bold.}
    \label{tab:non_zero_benchmark_appendix}
    \addtolength{\tabcolsep}{-0.29em}
    \begin{tabular}{lllllllllllllllll}
    \toprule
         & \multicolumn{6}{c}{Zero-shot} \\
    \cmidrule{2-7}
        Method & \multicolumn{1}{c}{Gecco} & \multicolumn{1}{c}{PSM} & \multicolumn{1}{c}{Dn} & 
         \multicolumn{1}{c}{Gen} & \multicolumn{1}{c}{SWaT} & \multicolumn{1}{c}{CC} & \multicolumn{1}{c}{GHL}& \multicolumn{1}{c}{OPP}& \multicolumn{1}{c}{SMAP}& \multicolumn{1}{c}{MSL}& \multicolumn{1}{c}{MITDB}& \multicolumn{1}{c}{SVDB}& \multicolumn{1}{c}{Exath}& \multicolumn{1}{c}{SMD}& \multicolumn{1}{c}{LTDB}& \multicolumn{1}{c}{TAO}\\
    \toprule
    % Mom.+AT & $03.27$ & $17.34$ & $\mathbf{09.60}$ & $01.45$ & $16.20$ & $05.00$ & $01.63$ & $02.34$ & $05.50$ & $10.39$ & $03.12$ & $07.06$ & $13.04$ & $05.37$ & $20.09$ & $83.87$ \\
    % Mom.+RAG$_0$ & $03.81$ & $18.55$ & $05.55$ & $01.18$ & $\mathbf{32.87}$ & $06.16$ & $\mathbf{04.28}$ & $02.29$ & $\mathbf{29.93}$ & $14.81$ & $05.09$ & $\mathbf{35.21}$ & $95.57$ & $09.35$ & $40.64$ & $\mathbf{88.26}$ \\
    Mom.+RAG$_3$ & $03.71$ & $27.66$ & $15.85$ & $02.07$ & $29.01$ & $01.96$ & $09.30$ & $05.41$ & $44.30$ & $19.46$ & $10.21$ & $36.08$ & $90.55$ & $12.62$ & $40.36$ & $16.92$ \\
    Mom.+RAG$_5$ & $04.46$ & $27.65$ & $15.88$ & $03.72$ & $30.66$ & $02.77$ & $09.50$ & $05.32$ & $44.53$ & $17.91$ & $10.02$ & $36.07$ & $91.17$ & $14.39$ & $38.79$ & $16.34$ \\
    Mom.+RAG$_7$ & $03.17$ & $27.66$ & $15.86$ & $02.20$ & $34.25$ & $00.91$ & $09.29$ & $04.88$ & $41.08$ & $20.15$ & $11.06$ & $36.56$ & $90.78$ & $14.43$ & $39.00$ & $16.23$ \\
    \midrule
    % TSP.+AT & $03.29$ & $17.42$ & $09.54$ & $01.40$ & $16.31$ & $05.00$ & $01.62$ & $02.33$ & $05.47$ & $10.32$ & $03.11$ & $07.03$ & $12.88$ & $05.35$ & $20.00$ & $83.82$ \\
    % TSP.+RAG$_0$ & $04.98$ & $18.00$ & $07.25$ & $01.01$ & $\mathbf{37.09}$ & $06.20$ & $\mathbf{06.04}$ & $03.75$ & $24.09$ & $17.14$ & $05.93$ & $\mathbf{28.94}$ & $92.21$ & $07.31$ & $34.12$ & $87.67$ \\
    TSP.+RAG$_3$ & $17.11$ & $28.15$ & $18.48$ & $01.07$ & $35.87$ & $00.58$ & $08.78$ & $07.62$ & $37.87$ & $21.67$ & $10.07$ & $28.24$ & $87.68$ & $11.05$ & $36.07$ & $16.15$ \\
    TSP.+RAG$_5$ & $15.44$ & $27.70$ & $27.77$ & $01.22$ & $36.76$ & $03.32$ & $09.23$ & $06.56$ & $41.94$ & $19.60$ & $11.41$ & $28.83$ & $87.70$ & $11.48$ & $33.65$ & $16.07$ \\
    TSP.+RAG$_7$ & $11.06$ & $27.89$ & $18.15$ & $1.03$ & $40.92$ & $00.48$ & $08.65$ & $04.77$ & $38.84$ & $20.42$ & $11.84$ & $27.38$ & $88.28$ & $11.27$ & $32.69$ & $16.31$ \\
    \midrule
    % TTM+AT & $03.68$ & $05.29$ & $\mathbf{09.98}$ & $01.40$ & $18.47$ & $04.96$ & $03.68$ & $02.37$ & $05.45$ & $10.28$ & $03.59$ & $06.88$ & $12.00$ & $05.29$ & $20.88$ & $85.09$ \\
    % TTM+RAG$_0$ & $05.84$ & $17.31$ & $05.26$ & $01.14$ & $68.77$ & $10.35$ & $\mathbf{04.86}$ & $04.75$ & $22.36$ & $19.3$ & $\mathbf{04.28}$ & $23.22$ & $92.79$ & $08.23$ & $33.76$ & $\mathbf{86.11}$ \\
    TTM+RAG$_3$ & $11.95$ & $27.67$ & $15.99$ & $01.01$ & $75.23$ & $03.9$ & $12.17$ & $17.09$ & $41.64$ & $21.26$ & $07.17$ & $29.83$ & $89.06$ & $15.10$ & $35.66$ & $16.06$ \\
    TTM+RAG$_5$ & $12.48$ & $27.78$ & $15.87$ & $07.14$ & $75.91$ & $02.08$ & $10.72$ & $13.77$ & $47.37$ & $25.34$ & $07.30$ & $29.72$ & $88.72$ & $15.80$ & $34.68$ & $16.16$ \\
    TTM+RAG$_7$ & $09.56$ & $27.65$ & $16.03$ & $05.30$ & $77.76$ & $04.36$ & $10.87$ & $09.05$ & $40.72$ & $20.36$ & $07.41$ & $29.58$ & $88.89$ & $16.75$ & $34.27$ & $16.10$ \\
    \bottomrule
    \end{tabular}
\end{table*}

\section*{Hyperparameters}
For all our experiments, we use the AdamW optimizer with a learning rate of $0.001$ and a cosine annealing scheduler. 
The value of $\lambda$ is set to $1$ and that of $\psi$ is set to $3$. 
We use a batch size of $256$ with number of epochs set to $50$ and $C=10$.
The context length (size of each window is set to $512$) with $H=16$.
\end{document}